\documentclass[10pt,twocolumn,letterpaper]{article}

\usepackage{cvpr}              

\usepackage{graphicx}
\usepackage{amsmath}
\usepackage{amssymb}
\usepackage{booktabs}
\usepackage{times}
\usepackage{epsfig}
\usepackage{graphicx}
\usepackage{amsmath}
\usepackage{amssymb}
\usepackage{multirow}
\usepackage{booktabs}
\usepackage{color}
\usepackage{colortbl}
\usepackage[table,xcdraw]{xcolor}
\usepackage[accsupp]{axessibility} 
\newcommand\blfootnote[1]{%
  \begingroup
  \renewcommand\thefootnote{}\footnote{#1}%
  \addtocounter{footnote}{-1}%
  \endgroup
}

\usepackage[pagebackref,breaklinks,colorlinks]{hyperref}

\usepackage[capitalize]{cleveref}
\crefname{section}{Sec.}{Secs.}
\Crefname{section}{Section}{Sections}
\Crefname{table}{Table}{Tables}
\crefname{table}{Tab.}{Tabs.}

\def\cvprPaperID{00006} 
\def\confName{CVPR}
\def\confYear{2022}

\begin{document}

\title{Semantically Grounded Visual Embeddings for Zero-Shot Learning}


\author{Shah Nawaz$^{1*\dagger}$, Jacopo Cavazza$^{1*}$, Alessio {Del Bue}$^{1,2}$ \\
$^{1}$Pattern Analysis \& Computer Vision (PAVIS) - Istituto Italiano di Tecnologia (IIT),\\ 
$^{2}$Visual Geometry \& Modelling (VGM) - Istituto Italiano di Tecnologia (IIT)
\\
{\tt\small \{shah.nawaz,jacopo.cavazza,alessio.delbue\}@iit.it}
}

\maketitle

\begin{abstract}
\blfootnote{*Equal contribution}
\blfootnote{$\dagger$ Current Affiliation: Deutsches Elektronen-Synchrotron (DESY)}
\blfootnote{Email: shah.nawaz@desy.de}
Zero-shot learning methods rely on fixed visual and semantic embeddings, extracted from independent vision and language models, both pre-trained for other large-scale tasks. This is a weakness of current zero-shot learning frameworks as such disjoint embeddings fail to adequately associate visual and textual information to their shared semantic content. Therefore, we propose to learn semantically grounded and enriched visual information by computing a joint image and text model with a two-stream network on a proxy task. To improve this alignment between image and textual representations, provided by attributes, we leverage ancillary captions to provide grounded semantic information. Our method, dubbed joint embeddings for zero-shot learning is evaluated on several benchmark datasets, improving the performance of existing state-of-the-art methods in both standard ($+1.6$\% on aPY, $+2.6\%$ on FLO) and generalized ($+2.1\%$ on AWA$2$, $+2.2\%$ on CUB) zero-shot recognition.
\end{abstract}

Zero-shot learning (ZSL) aims at classifying images into new ``unseen'' categories at test time without having been provided any corresponding visual examples in the training stage. This is possible by taking advantage of \textit{semantic embeddings} describing the visual categories by means of auxiliary text information i.e. attributes. In recent years, several approaches have been proposed to tackle ZSL: A metric/score can be learnt to quantify the compatibility between semantic embeddings and pre-computed visual features (\emph{visual embedding})~\cite{xian2018tPAMI}. Alternatively, visual features can be synthesized from semantic embeddings with adversarial generative training \cite{narayan2020latent,xian2018feature,ye2017zero}. In these terms, ZSL and its extension named generalised ZSL (GZSL)~\cite{xian2018tPAMI}, naturally appear as multimodal classification problems. However, visual and semantic embeddings adopted in  ZSL approaches are usually extracted \emph{independently} from ancillary tasks. In fact, visual embedding are usually features trained for object/scene recognition on classical vision benchmarks such as ImageNet~\cite{he2016deep,simonyan2014very}. On the other hand, semantic embeddings in ZSL are either manually defined, by collecting attributes which describe a certain category, or extracted using natural language processing~\cite{devlin2018bert,he2016deep,mikolov2013distributed,simonyan2014very}.

\begin{figure}[t!]
    \centering
    \includegraphics[width=\columnwidth]{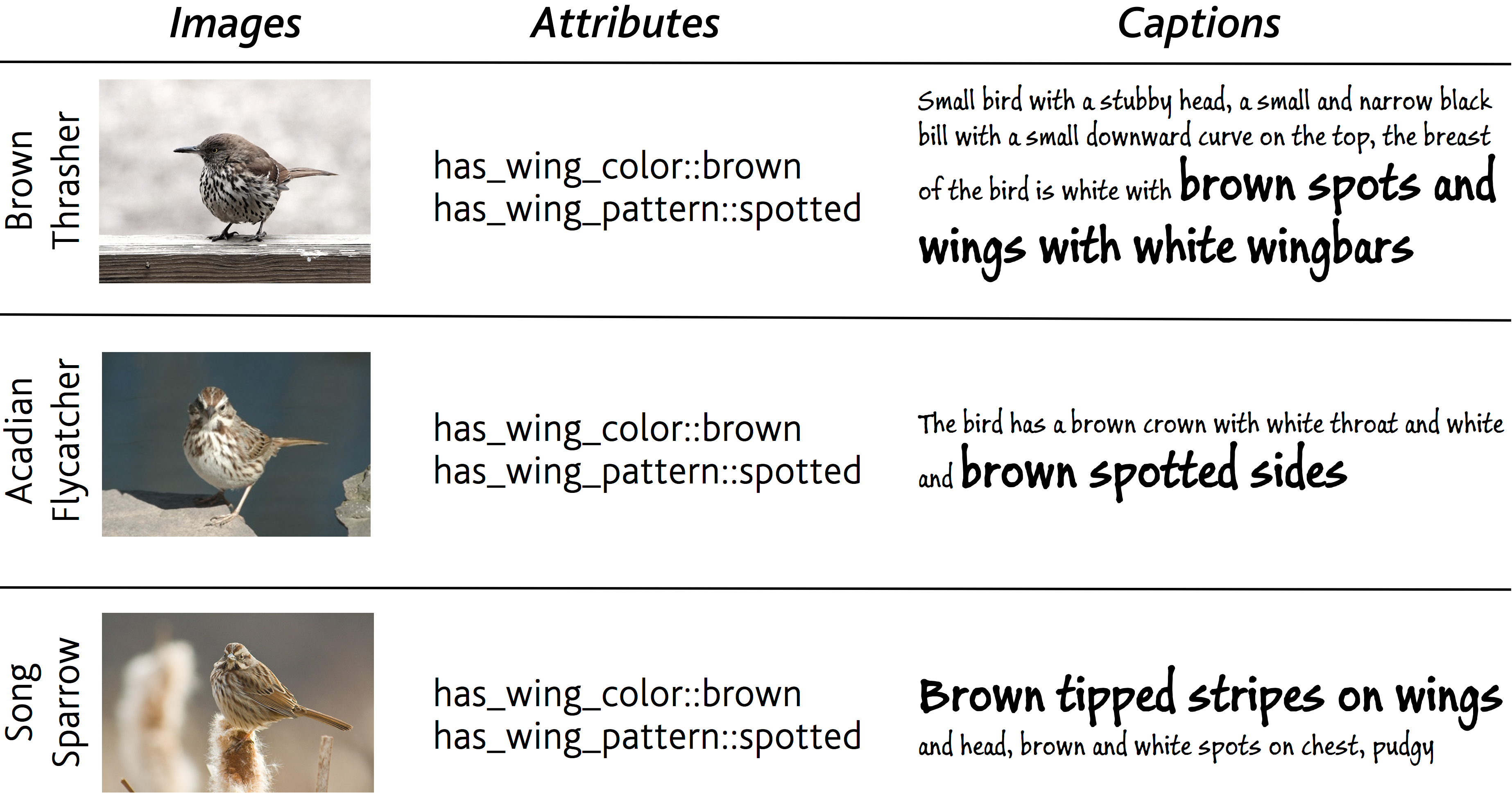}
    \caption{For these three bird species of CUB \cite{CUB} dataset, the attributes ``brown spotted wings'' is not discriminative. In fact, a ``Brown Thrasher'' is characterized by the presence of white wingbars. An ``Arcadian Flycatcher'' features spots which are present over the whole side of the animal -- not just over the wings. A ``Song Sparrow'' shows peculiar tips over the stripes.
    We posit that captions can provide ancillary semantic information which we propose to exploit altogether with attributes and visual data. 
    }
    \label{fig:motivational}
\end{figure}

Independently extracted semantic embeddings convey only \emph{general} information about a given visual category. For this reason, their sensitivity may be limited in describing \emph{specific} semantic nuances that actually help in discriminating between another category whose instances look extremely similar (see Figure \ref{fig:motivational}). In this paper, we posit that this problem is affecting zero-shot learning and, in order to mitigate it,
we propose to exploit captions to inject semantic information into visual embeddings. This is achieved by means of a joint training, finalized at better enriching the embeddings using both visual and semantic cues, in order to ease a zero-shot learning recognition paradigm.



Generally, a jointly learned image and text representation is referred as  \textit{pretrain-then-transfer}, a defacto standard paradigm to process vision and language information~\cite{baltrusaitis2019multimodal,lu2019vilbert}
The success of \textit{pretrain-then-transfer} learning is instrumental in developing methods which can capture connections between vision and language domains on so-called ``proxy tasks'' with large-scale image and text datasets i.e. MSCOCO~\cite{lin2014microsoft}, Flickr30K~\cite{plummer2015flickr30k}, Conceptional Captions (ConcCapt)~\cite{sharma2018conceptual}, Visual Genome~\cite{krishna2017visual}. Therefore, in the recent years, an increased interest was registered in unifying models as to learn joint image and text representations~\cite{chen2020uniter,lu2019vilbert,qi2020imagebert,tan2019lxmert}. These models utilize cross-modal attention mechanisms to capture interaction between image regions and fine-grained text information. 
However, zero-shot learning methods cannot leverage these powerful models to improve visual or semantic information because the zero-shot learning problem does not account for fine-grained text description for visual information. Therefore, we solve for this deficiency by presenting a framework to learn joint image and text representation to improve visual information for ZSL and GZSL tasks. 

We propose a novel framework, called joint-embeddings for zero-shot learning (\textbf{JE-ZSL}), to learn a semantically-grounded visual representation. We use a two-stream network, consisting of two layers of non-linearities on top of the image and captions, trained with a bi-directional loss function. We implement structure-preserving constraints to align the alternative visual and textual sources of information.
We posit that, by means of our joint training, we can better capture the interaction between semantically-related image and text components, ultimately bridging the gap between visual and semantic representations and, in turn, improving the performance of a zero-shot learning mechanism. 
We evaluate our proposed semantically grounded embeddings on various well-know compatibility function and GAN based  methods, improving standard and generalised zero-shot learning performance on coarse-grained and fine-grained dataset. This empirically demonstrates the benefits of corroborating the representation adopted for visual data in zero-shot learning by means of semantically grounded information. 


\begin{figure*}[!t]
 \centering
 \includegraphics[width=0.9\textwidth]{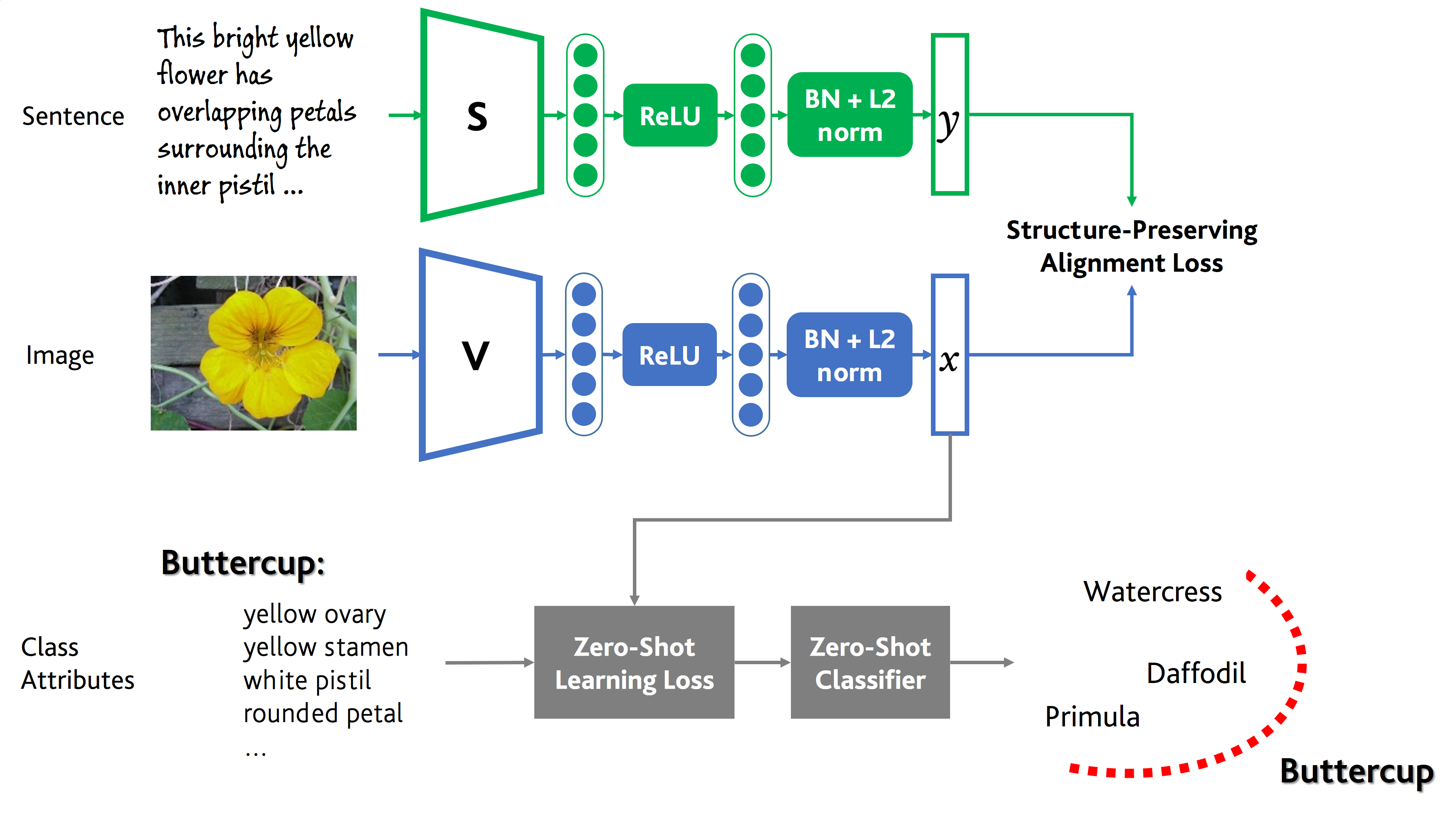}
      \caption{\textit{Overview of Joint Embeddings for Zero-Shot Learning} (\textbf{JE-ZSL}). Our model consists of a visual and semantic embedding modules, $\mathbf{V}$ and $\mathbf{S}$ respectively, which are responsible for the joint training of the image embeddings starting from textual cues (here, sentences). We stack two fully-connected layers, using ReLU non-linearity, batch normalization (BN), L2 normalization and a Structure-Preserving Alignment Loss which is responsible of injecting semantic patterns into visual descriptors (see Section \ref{sez:spal}). Once the jointly trained visual embeddings $\mathbf{x}$ are optimized over a set of seen classes (in the figure, watercress, daffodil, primula, ...), they can be used to train the optimization objective of an arbitrary zero-shot learning framework. Consequently, even if we never saw an image of a ``buttercup'', we can still manage to classify it (decision boundary represented in red) while leveraging the side information which provides the attributes that the class ``buttercup'' is expected to own.}
  \label{fig:arch}
\end{figure*}

\section{Related Work}
We summarize previous work on both joint image and text representations (Section \ref{sez:rw1}) along with standard/generalized zero-shot learning (Section \ref{sez:rw2}).

\subsection{Joint Image and Text Representation}\label{sez:rw1}
In the recent years, several works have been proposed to learn joint image and text representation for various vision and languages tasks in a multimodal setting~\cite{baltrusaitis2019multimodal,arshad2019aiding,nawaz2019cross,gallo2017multimodal,faghri2017vse++}.
Usually, joint image and text representation strategies start with separate vision and language models (pre-trained for other large-scale tasks) and then perform a task-specific training~\cite{antol2015vqa,faghri2017vse++}. However, the last years have seen a paradigm shift in learning joint image and text representations on proxy tasks designed to bridge the semantic gap between visual and textual clues leveraging large-scale datasets~\cite{chen2020uniter,li2019visualbert,lu2019vilbert,qi2020imagebert, tan2019lxmert}. These joint representations transfer to various downstream vision and language tasks including visual question answering, grounding referring expression, visual commonsense reasoning and image retrieval, without the necessity of extra-training. In this spirit, we proposed to learn semantically grounded enriched visual embedding using joint image and text representation on a proxy task. The key differences between our JE-ZSL approach and the other methods are twofold: (\textit{i}) Interaction between image and text representations only occur at the last layer of the two-stream network; and (\textit{ii}) JE-ZSL uses global visual and text representation to capture interaction between image and textual clues. These key differences enable our JE-ZSL model to handle downstream zero-shot learning tasks.

\subsection{Zero-shot Learning}\label{sez:rw2}

In ZSL, a compatibility function 
is used to measure the matching between visual and semantic embeddings. For that scope, either metric  \cite{jiang2018learning,kodirov2017semantic,ESZSL,zhang2018zero}, scoring \cite{li2018discriminative,song2018transductive,GFZSL}  or ranking functions \cite{changpinyo2018classifier,guo2019dual} are adopted. 
However, the main drawback of those methods is the shallowness of the architectural design. As a result, an improvement in the recognition of unseen classes usually corresponds to a decreased performance in the seen classes and vice-versa~\cite{xian2018tPAMI}, damaging the overall generalization capabilities. 
In this paper, we propose to tackle this issue by the usage of joint embeddings, which we conjecture that can help shallow models in better generalizing.

As an alternative strategy to compatibility functions, the pivotal paper \cite{Xian_2018_CVPR} proposed to directly synthesize visual descriptors conditioned on semantic embeddings. A Generative Adversarial Network (GAN) is trained to synthesize visual features which are indistinguishable from the ones extracted from a ResNet$-101$ backbone. At the deployment stage, the very same model can be utilized to generate synthetic visual features for the unseen classes. Finally, a shallow softmax classifier is trained using both real features from seen classes and unseen features from the unseen ones. This baseline architectural design was recently advanced in various approaches, such as the adoption of variational inference to boost the generator \cite{huang2019generative,narayan20,Schonfeld_2019_CVPR,Xian_2019_CVPR,Zhu:ICCV19}. Additional loss functions were investigated to better handle the problem, like cycle-consistency  \cite{felix2018multi}, triplet \cite{cacheux2019modeling} or contrastive loss \cite{Jiang:ICCV19}. In this paper, we show that the semantic conditioning used to generate visual features can be furthermore boosted by the usage of joint embeddings: since they are endowed of semantically-grounded visual information, the ``gap'' between semantic and visual embeddings is already bridged at the input level already.

\section{Joint Embeddings for Zero-Shot Learning: Methodology}

Our proposed strategy JE-ZSL is to ground visual embeddings through the usage of textual information for the sake of enhancing zero-shot learning methods from the data representation point of view.

\subsection{Overview of the Proposed Architecture}\label{sez:arch}

In Figure~\ref{fig:arch}, we visualize the joint embeddings model that we propose. It is composed by three input streams, each of them corresponding to a different type of input data (images, sentences and class attributes). 
The modules $\mathbf{V}$ and $\mathbf{S}$ provide vectorial representation extracted from ResNet$-101$~\cite{he2016deep} and Bidirectional Encoder Representations from Transformers (BERT)~\cite{devlin2018bert} models respectively to learn joint embeddings. 
On top of the output of $\mathbf{V}$ and $\mathbf{S}$, JE-ZSL stacks two-fully connected layers, the first of which followed by a rectified linear unit (ReLU) and the second of which is followed by batch-norm operator followed by an L2 normalization. These two fully-connected layers are then responsible for the joint training of the visual embeddings $x$ supported by the sentence embeddings $y$, by means of a structure-preserving alignment loss~\cite{wang2016learning} (see Section \ref{sez:spal}). Once visual embeddings are optimized, they can be used in an arbitrary zero-shot learning framework, together with attributes - or other class embeddings - without requiring any modification in the pipeline of the zero-shot learner, thus ensuring a broad applicability (additional details in Section \ref{sez:zsl}).

\subsection{Structure-Preserving Alignment Loss}\label{sez:spal}

The JE-ZSL loss function is defined as follow. Given a visual embedding $x_i$, let  $Y_i^+$  and $Y_i^-$ represent sets of positive and negative sentence embeddings respectively. 
The distance between a positive pair should be smaller than the distance between a negative pair with margin $m$:
\begin{equation}
\label{eq:sent-pos}
d\left(x_{i}, y_{j}\right)+m<d\left(x_{i}, y_{k}\right) \quad \forall y_{j} \in Y_{i}^{+}, \forall y_{k} \in Y_{i}^{-},
\end{equation}
where $x_i$ and  $y_j$ represents positive pair while $x_i$ and $y_k$ is negative pair.
Similarly, given a sentence $y_{i^{\prime}}$:
\begin{equation}
d\left(x_{j^{\prime}}, y_{i^{\prime}}\right)+m<d\left(x_{k^{\prime}}, y_{i^{\prime}}\right) \quad \forall f_{j^{\prime}} \in X_{i^{\prime}}^{+}, \forall x_{k^{\prime}} \in X_{i^{\prime}}^{-},
\end{equation}
where $X_{i^{\prime}}^{+}$ and $X_{i^{\prime}}^{-}$ represents the sets of positive and negative images for  $y_{i^{\prime}}$.

In addition, the loss function includes neighborhood constraints to project semantically similar images and sentences close to each other in the embedding space. Let $N(x_i)$ represent the neighborhood of $x_i$ with images that contains the same concept. In our scenario, it contains images represented by the set of sentences associated with $x_i$ 
Furthermore, a margin $m$ is enforced between $N(x_i)$ and any point outside of the neighborhood:

\begin{equation}
\label{eq:n1}
d\left(x_{i}, x_{j}\right)+m<d\left(x_{i}, x_{k}\right) \quad \forall x_{j} \in N\left(x_{i}\right), \forall x_{k} \notin N\left(x_{i}\right)\end{equation}

Similarly to Eq.~\ref{eq:n1}, the constraint is enforced for sentences: 
\begin{equation}d\left(y_{i^{\prime}}, y_{j^{\prime}}\right)+m<d\left(y_{i^{\prime}}, y_{k^{\prime}}\right) \quad \forall y_{j^{\prime}} \in N\left(y_{i^{\prime}}\right), \forall y_{k^{\prime}} \notin N\left(y_{i^{\prime}}\right)\end{equation}

Finally, constraints are converted to the training objective using hinge loss. The resulting loss function is given by:

$$\begin{aligned}
L(X, Y) &=\sum_{i, j, k} \max \left[0, m+d\left(x_{i}, y_{j}\right)-d\left(x_{i}, y_{k}\right)\right] \\
&+\lambda_{1} \sum_{i^{\prime}, j^{\prime}, k^{\prime}} \max \left[0, m+d\left(x_{j^{\prime}}, y_{i^{\prime}}\right)-d\left(x_{k^{\prime}}, y_{i^{\prime}}\right)\right] \\
&+\lambda_{2} \sum_{i, j, k} \max \left[0, m+d\left(x_{i}, x_{j}\right)-d\left(x_{i}, x_{k}\right)\right] \\
&+\lambda_{3} \sum_{i^{\prime}, j^{\prime}, k^{\prime}} \max \left[0, m+d\left(y_{i^{\prime}}, y_{j^{\prime}}\right)-d\left(y_{i^{\prime}}, y_{k^{\prime}}\right)\right].
\end{aligned}$$

The hyperparameter $\lambda_{1}$ balances strengths of both ranking terms and the value is fixed to $2$. Similarly, $\lambda_{2}$ and $\lambda_{3}$ controls the neighborhood constraint and values are set to $0.1$ or $0.2$ respectively~\cite{wang2016learning}. The distance $d$ is fixed to be the Euclidean distance. 
In addition, triplets are selected within the mini-batch only.
To learn our joint embeddings, the network is trained in end-to-end fashion with large-scale datasets MSCOCO~\cite{lin2014microsoft}, Flickr30K~\cite{plummer2015flickr30k}, Conceptional Captions (ConcCapt))~\cite{sharma2018conceptual}, using captioning annotations to leverage the fine-grained datasets Caltech-UCSD-Birds (CUB)~\cite{welinder2010caltech} and Oxford Flowers (FLO)~\cite{nilsback2008automated}.

\subsection{Joint Embeddings for Zero-shot Learning}\label{sez:zsl}

Once the joint embeddings $x$ are extracted by means of the structure-preserving alignment loss, they are then fed to a second computational module which is responsible for the zero-shot learning task. In order to classify a pool of unseen classes $c \in \mathcal{C}_U$, we take advantage of training data $(x_i,c_i), i = 1,\dots,N$ where the labels $c_i$ belong to the seen classes ($c_i \in \mathcal{C}_S)$ for every $i$. In order to bridge the gap between seen and unseen classes, either of them need to be describes in terms of class-embeddings $a = a_c$ such as manually defined attributes. The learning problem writes as minimizing the empirical risk defined through the following loss function
\begin{equation}\label{eq:generic-zsl}
   \sum_{i = 1}^N\mathcal{L}_{\rm ZSL}(x_i, a_{c_i}; \theta),
\end{equation}
which depends upon the learnable parameters $\theta$. In turn, the parameters $\theta$ can be optimized exploiting different strategies, using alternative expressions for $\mathcal{L}_{\rm ZSL}$ in eq \eqref{eq:generic-zsl}. For methods based on a compatibility function, $\mathcal{L}_{\rm ZSL}$ is chosen to be either a distance \cite{jiang2018learning,kodirov2017semantic,ESZSL,zhang2018zero}, scoring \cite{li2018discriminative,song2018transductive,GFZSL}  or ranking function \cite{changpinyo2018classifier,guo2019dual}. The optimization attempts to match $x_i$ and $a_i$ when and only when they belongs to the same class. Additional regularization terms are used to make sure that the distance/scoring/ranking function $\mathcal{L}_{\rm ZSL}$ can generalize towards unseen classes. In fact, when a new test instance $\tilde{x}$ occurs, inference is solved as
\begin{equation}\label{eq:generic-inference}
    \arg \max_c F_{\rm ZSL}(\tilde{x},\theta),
\end{equation}
being $F_{\rm ZSL}$ a 1-nearest neighbor classifier based on the learned compatibility function $\mathcal{L}_{\rm ZSL}$. Alternatively, $F_{\rm ZSL}$ can be chosen to be a softmax classifier trained on 
real joint embeddings from seen classes $\mathcal{C}_S$ and synthesized features from the unseen classes $\mathcal{C}_U$. For the feature synthesis, models such as generative adversarial networks \cite{Atzmon19,Felix18,Huang19,Xian_2018_CVPR,Yu20} or variational autoencoders \cite{Gao20,narayan20,Schonfeld19,Xian_2019_CVPR} are adopted to generate visual descriptors conditioned on the class-specific embeddings $a_c$, $c \in \mathcal{C}_U$.

\section{Experiments}
\subsection{Data Sources}
{\bf Datasets for Joint Embeddings.} We trained joint image and text model on three publicly available datasets including MSCOCO~\cite{lin2014microsoft}, Flickr30K~\cite{plummer2015flickr30k} and Conceptional Captions (ConcCapt)~\cite{sharma2018conceptual}. MSCOCO dataset contains 123,287 images, and each image is annotated with five text descriptions or captions. Similarly, Flickr30K contains 31,783 images along with five captions for each image. While, ConcCapt dataset consists of $\sim\!\!3.3$ million image and text pairs collected from alt-text enabled images on the web.
At time of access, some links are broken which resulted in $\sim\!\!1.3$ million image and text pairs.

{\bf Datasets for Zero-shot Learning.} We evaluate JE-ZSL on two fine-grained and two coarse-grained zero-shot object recognition data sources: CUB~\cite{welinder2010caltech}, FLO~\cite{nilsback2008automated}, Animals with Attributes (AWA$2$)~\cite{xian2018zero} and Attribute Pascal \& Yahoo (aPY)~\cite{farhadi2009describing} containing $200$, $102$, $717$, $50$ and $32$ categories, respectively.  For fair comparison, we followed the standard zero-shot splits proposed by Xian et al.~\cite{Xian_2018_CVPR}.

\subsection{Backbone ZSL Methods}
We evaluate the performance of JE-ZSL when adopting the following backbone zero-shot learning methods: We select one of the most efficient methods based on a compatibility function~\cite{Jiang18}, while also including two state-of-the-art approaches based on feature generation~\cite{narayan20,xian2018feature}. Further details on these methods are provided below, including how we combined them with our joint embeddings. 

Coupled Dictionary Learning~\cite{Jiang18} (\textbf{CDL}) is composed of the following stages. First, given the semantic embeddings of the classes to be recognized, visual prototypes are learnt by preserving neighborhood relationship across the visual and semantic spaces. This is obtained through a structure alignment module and implemented in the form of visual and semantic dictionaries which are then learnt and coupled. We take advantage of our jointly training embeddings for the sake of learning these dictionary-based representations.

\textbf{f-CLSWGAN~\cite{xian2018feature}} 
is implemented as a Wasserstein GAN which attempts to generate visual features, which are similar to the ones available for seen classes, while also performing semantic-to-visual translation. To better enhance the discrimination capability of the learnt feature, a classification loss is used. Instead of standard ResNet$-101$ descriptors, we take advantage of jointly training embedding as the real features, supervising the adversarial learning. Once features have been generated for the unseen classes, a softmax classifier is trained from real joint embeddings from the seen classes and synthesized features for the unseen ones. The trained softmax classifier is the model used for the final inference.

A recent generative approach, called \textbf{tf-VAEGAN}~\cite{narayan20}, improved upon~\cite{xian2018feature} under multiple perspectives. First, the generation capability of a GAN are paired with a variational autoencoder and a cycle consistency loss is adopted, aiming at predicting attributes from synthesized features using a regression paradigm. We modify the original implementation by exploiting joint embeddings as either the real features from the seen classes and the input data for the approach regressing attributes. For the final inference stage, the concatenation of the generated visual features and regressed semantic embeddings are used to train the final softmax classifier. Joint embeddings are adopted also at this stage.

\subsection{Implementation Details}
We used listed hyperparameters to learn joint image and text embedding with a structure perserving alignment loss~\cite{wang2016learning}. Similarly, we employed listed hyperparameters for zero-shot learning methods~\cite{Jiang18,narayan2020latent,xian2018feature}. \\
{\bf Baseline Visual and Semantic Embeddings.} For visual embedding, we extract $2048$-dim top layer pooling units of ResNet$-101$~\cite{he2016deep}.  For the semantic embeddings, we employ the class-level attributes for CUB ($312-d$), AWA ($85-d$) and aPY ($64-d$). For FLO, fine-grained visual descriptions of image with size ($1024-d$) are extracted from a character-based CNN-RNN~\cite{reed2016learning}.\\
{\bf Semantically Grounded Embeddings.} The proposed embedding is extracted from joint image and text representation trained with structure preserving alignment loss. 
For fair comparison, we learn joint image and text representation with ResNet$-101$ embeddings. \\
%
{\bf Evaluation Metrics.} To evaluate the performance on standard zero-shot learning, we utilized the recommended top-1 classification accuracy over seen classes (T1). Similarly, we adopted common metrics for generalized zero-shot learning as well \cite{xian2018tPAMI}: Mean per-class accuracy over seen and unseen classes $s$ and $u$, respectively, while reporting also their harmonic mean $H$.

\section{Joint Embeddings for Zero-Shot Learning: Ablation Study and the Fine-Grained case}
In this section, we provide an experimental analysis to understand and evaluate the impact of two distinct factors influencing the performance of JE-ZSL. First, we present an ablation study with to monitor if there is an effect of varying the dimensionality of our proposed joint embeddings while training them using the structure-preserving alignment loss (presented in Section \ref{sez:spal}) on various image-text datasets. Then, we will evaluate the impact of using fine-grained captions leveraging fine-grained datasets.

\begin{table*}[]
\centering
\begin{tabular}{|lll|cccc|cccc|}
\hline
                              &                             & \textit{}     & \multicolumn{4}{c|}{AWA$2$ \cite{AWA} }                                                                                                                                                                                                                                                       & \multicolumn{4}{c|}{aPY \cite{APY} }                                                                                                                                                                                                                                   \\
                              &                             & \textit{ }  & T1                                                           & $u$                                                            & $s$                                                            & \multicolumn{1}{c|}{$H$}                                                            & T1                                                           & $u$                                                            & $s$                                                            & $H$                                                            \\ \hline \hline
\multicolumn{3}{|r|}{\textit{Baseline} $\colon$ \textit{CDL} \cite{Jiang18}} & \textit{69.9}                                                & \textit{28.1}                                                & \textit{73.5}                                                & \multicolumn{1}{c|}{\textit{40.6}}                                                & \textit{43.0}                                                & \textit{19.8}                                                & \textit{48.6}                                                & \textit{28.1}                                                \\ \cline{2-11} 
                              &                             & \scalebox{0.7}{$d = 256$}  & 65.6                                                         & 24.2                                                         & 71.7                                                         & \multicolumn{1}{c|}{36.2}                                                         & \cellcolor[HTML]{CBCEFB}{\color[HTML]{330001} \textbf{44.1}} & 18.1                                                         & 46.1                                                         & 26.0                                                         \\
                              &                             & \scalebox{0.7}{$d = 512$}  & 67.1                                                         & 26.2                                                         & \cellcolor[HTML]{CBCEFB}{\color[HTML]{330001} \textbf{76.1}} & \multicolumn{1}{c|}{39.0}                                                         & \cellcolor[HTML]{CBCEFB}{\color[HTML]{330001} \textbf{44.8}} & 19.4                                                         & 47.3                                                         & 27.5                                                         \\
                              &                             & \scalebox{0.7}{$d = 1024$} & 69.2                                                         & 27.1                                                         & \cellcolor[HTML]{CBCEFB}{\color[HTML]{330001} \textbf{74.6}} & \multicolumn{1}{c|}{39.8}                                                         & \cellcolor[HTML]{CBCEFB}{\color[HTML]{330001} \textbf{45.6}} & \cellcolor[HTML]{CBCEFB}{\color[HTML]{330001} \textbf{20.1}} & \cellcolor[HTML]{CBCEFB}{\color[HTML]{330001} \textbf{49.8}} & \cellcolor[HTML]{CBCEFB}{\color[HTML]{330001} \textbf{28.6}} \\
\multirow{-4}{*}{\textbf{JE-ZSL} \textit{(ours)}} & \multirow{-4}{*}{Flickr30K} & \scalebox{0.7}{$d = 2048$} & 68.2                                                         & 27.9                                                         & \cellcolor[HTML]{CBCEFB}{\color[HTML]{330001} \textbf{77.1}} & \multicolumn{1}{c|}{\cellcolor[HTML]{CBCEFB}{\color[HTML]{330001} \textbf{41.0}}} & \cellcolor[HTML]{CBCEFB}{\color[HTML]{330001} \textbf{45.2}} & \cellcolor[HTML]{CBCEFB}{\color[HTML]{330001} \textbf{22.3}} & \cellcolor[HTML]{CBCEFB}{\color[HTML]{330001} \textbf{50.4}} & \cellcolor[HTML]{CBCEFB}{\color[HTML]{330001} \textbf{30.9}} \\ \cline{2-11} 
                              &                             & \scalebox{0.7}{$d = 256$}  & 68.1                                                         & 27.1                                                         & 69.3                                                         & \multicolumn{1}{c|}{39.0}                                                         & \cellcolor[HTML]{CBCEFB}\textbf{45.1}                        & 18.6                                                         & 48.6                                                         & \cellcolor[HTML]{CBCEFB}\textbf{26.9}                        \\
                              &                             & \scalebox{0.7}{$d = 512$}  & {\color[HTML]{000000} 69.3}                                  & \cellcolor[HTML]{CBCEFB}{\color[HTML]{000000} \textbf{28.4}} & \cellcolor[HTML]{CBCEFB}{\color[HTML]{000000} \textbf{75.6}} & \multicolumn{1}{c|}{\cellcolor[HTML]{CBCEFB}{\color[HTML]{000000} \textbf{41.3}}} & \cellcolor[HTML]{CBCEFB}{\color[HTML]{000000} \textbf{46.3}} & \cellcolor[HTML]{CBCEFB}{\color[HTML]{000000} \textbf{23.2}} & \cellcolor[HTML]{CBCEFB}{\color[HTML]{000000} \textbf{50.2}} & \cellcolor[HTML]{CBCEFB}{\color[HTML]{000000} \textbf{31.7}} \\
                              &                             & \scalebox{0.7}{$d = 1024$} & \cellcolor[HTML]{CBCEFB}{\color[HTML]{000000} \textbf{69.5}} & \cellcolor[HTML]{CBCEFB}{\color[HTML]{000000} \textbf{29.0}} & \cellcolor[HTML]{CBCEFB}{\color[HTML]{000000} \textbf{75.6}} & \multicolumn{1}{c|}{\cellcolor[HTML]{CBCEFB}{\color[HTML]{000000} \textbf{41.9}}} & \cellcolor[HTML]{CBCEFB}{\color[HTML]{000000} \textbf{45.3}} & \cellcolor[HTML]{CBCEFB}{\color[HTML]{000000} \textbf{20.4}} & \cellcolor[HTML]{CBCEFB}{\color[HTML]{000000} \textbf{49.9}} & \cellcolor[HTML]{CBCEFB}{\color[HTML]{000000} \textbf{28.7}} \\
\multirow{-4}{*}{\textbf{JE-ZSL} \textit{(ours)}}   & \multirow{-4}{*}{MSCOCO}    & \scalebox{0.7}{$d = 2048$} & \cellcolor[HTML]{CBCEFB}{\color[HTML]{000000} \textbf{70.1}} & \cellcolor[HTML]{CBCEFB}{\color[HTML]{000000} \textbf{28.4}} & \cellcolor[HTML]{CBCEFB}{\color[HTML]{000000} \textbf{72.1}} & \multicolumn{1}{c|}{\cellcolor[HTML]{CBCEFB}{\color[HTML]{000000} \textbf{40.7}}} & \cellcolor[HTML]{CBCEFB}{\color[HTML]{000000} \textbf{44.1}} & \cellcolor[HTML]{CBCEFB}{\color[HTML]{000000} \textbf{19.4}} & \cellcolor[HTML]{CBCEFB}{\color[HTML]{000000} \textbf{49.0}} & {\color[HTML]{000000} 27.8}                                  \\ \cline{2-11} 
                              &                             & \scalebox{0.7}{$d = 256$}  & 68.3                                                         & 26.1                                                         & 71.7                                                         & \multicolumn{1}{c|}{38.3}                                                         & 42.1                                                         & 20.2                                                         & 46.7                                                         & \cellcolor[HTML]{CBCEFB}\textbf{28.2}                        \\
                              &                             & \scalebox{0.7}{$d = 512$}  & 69.9                                                         & 28.1                                                         & 73.5                                                         & \multicolumn{1}{c|}{\cellcolor[HTML]{CBCEFB}\textbf{40.7}}                        & \cellcolor[HTML]{CBCEFB}\textbf{43.3}                        & \cellcolor[HTML]{CBCEFB}\textbf{21.8}                        & \cellcolor[HTML]{CBCEFB}\textbf{49.2}                        & \cellcolor[HTML]{CBCEFB}\textbf{30.2}                        \\
                              &                             & \scalebox{0.7}{$d = 1024$} & 69.3                                                         & 28.1                                                         & 77.1                                                         & \multicolumn{1}{c|}{\cellcolor[HTML]{CBCEFB}\textbf{41.2}}                        & \cellcolor[HTML]{CBCEFB}\textbf{44.8}                        & \cellcolor[HTML]{CBCEFB}\textbf{22.8}                        & \cellcolor[HTML]{CBCEFB}\textbf{50.4}                        & \cellcolor[HTML]{CBCEFB}\textbf{31.4}                        \\
\multirow{-4}{*}{\textbf{JE-ZSL} \textit{(ours)}}   & \multirow{-4}{*}{ConcCapt}  & \scalebox{0.7}{$d = 2048$} & \cellcolor[HTML]{CBCEFB}\textbf{71.0}                        & \cellcolor[HTML]{CBCEFB}\textbf{29.3}                        & 77.1                                                         & \multicolumn{1}{c|}{\cellcolor[HTML]{CBCEFB}\textbf{42.5}}                        & \cellcolor[HTML]{CBCEFB}\textbf{45.9}                        & \cellcolor[HTML]{CBCEFB}\textbf{24.5}                        & \cellcolor[HTML]{CBCEFB}\textbf{51.1}                        & \cellcolor[HTML]{CBCEFB}\textbf{33.1}                        \\ \hline \hline
\multicolumn{3}{|r|}{\textit{Baseline} $\colon$ \textit{f-CLSWGAN} \cite{xian2018feature} }  & \textit{68.2}                                                & \textit{57.9}                                                & \textit{61.4}                                                & \multicolumn{1}{c|}{\textit{59.6}}                                                & \textit{40.5}                                                & \textit{25.8}                                                & \textit{59.5}                                                & \textit{36.0}                                                \\ \cline{2-11} 
                              &                             & \scalebox{0.7}{$d = 256$}  & 67.0                                                         & 55.4                                                         & 58.2                                                         & \multicolumn{1}{c|}{56.8}                                                         & 38.6                                                         & \cellcolor[HTML]{CBCEFB}\textbf{28.1}                        & 18.0                                                         & 22.0                                                         \\
                              &                             & \scalebox{0.7}{$d = 512$}  & 65.6                                                         & \cellcolor[HTML]{CBCEFB}\textbf{59.0}                        & \cellcolor[HTML]{CBCEFB}\textbf{62.1}                        & \multicolumn{1}{c|}{\cellcolor[HTML]{CBCEFB}\textbf{60.5}}                        & \cellcolor[HTML]{CBCEFB}\textbf{39.3}                        & 24.5                                                         & 50.6                                                         & 33.0                                                         \\
                              &                             & \scalebox{0.7}{$d = 1024$} & 68.0                                                         & 55.2                                                         & \cellcolor[HTML]{CBCEFB}\textbf{65.8}                        & \multicolumn{1}{c|}{\cellcolor[HTML]{CBCEFB}\textbf{60.1}}                        & \cellcolor[HTML]{CBCEFB}\textbf{38.7}                        & \cellcolor[HTML]{CBCEFB}\textbf{27.9}                        & 44.9                                                         & 34.4                                                         \\
\multirow{-4}{*}{\textbf{JE-ZSL} \textit{(ours)}} & \multirow{-4}{*}{Flickr30K} & \scalebox{0.7}{$d = 2048$} & 65.0                                                         & 55.8                                                         & \cellcolor[HTML]{CBCEFB}\textbf{67.6}                        & \multicolumn{1}{c|}{\cellcolor[HTML]{CBCEFB}\textbf{61.1}}                        & \cellcolor[HTML]{CBCEFB}\textbf{37.6}                        & 23.4                                                         & 63.2                                                         & 34.1                                                         \\ \cline{2-11} 
                              &                             & \scalebox{0.7}{$d = 256$}  & \cellcolor[HTML]{CBCEFB}\textbf{72.7}                        & \cellcolor[HTML]{CBCEFB}\textbf{61.5}                        & \cellcolor[HTML]{CBCEFB}\textbf{62.3}                        & \multicolumn{1}{c|}{\cellcolor[HTML]{CBCEFB}\textbf{61.9}}                        & 39.3                                                         & \cellcolor[HTML]{CBCEFB}\textbf{31.4}                        & 27.2                                                         & \multicolumn{1}{c|}{29.2}                                    \\
                              &                             & \scalebox{0.7}{$d = 512$}  & \cellcolor[HTML]{CBCEFB}\textbf{70.7}                        & \cellcolor[HTML]{CBCEFB}\textbf{63.4}                        & \cellcolor[HTML]{CBCEFB}\textbf{64.0}                        & \multicolumn{1}{c|}{\cellcolor[HTML]{CBCEFB}\textbf{63.8}}                        & \cellcolor[HTML]{CBCEFB}\textbf{41.7}                        & \cellcolor[HTML]{CBCEFB}\textbf{26.5}                        & 48.5                                                         & \multicolumn{1}{c|}{34.3}                                    \\
                              &                             & \scalebox{0.7}{$d = 1024$} & \cellcolor[HTML]{CBCEFB}\textbf{70.4}                        & \cellcolor[HTML]{CBCEFB}\textbf{62.8}                        & \cellcolor[HTML]{CBCEFB}\textbf{63.8}                        & \multicolumn{1}{c|}{\cellcolor[HTML]{CBCEFB}\textbf{63.3}}                        & 38.0                                                         & 25.7                                                         & 49.9                                                         & \multicolumn{1}{c|}{34.0}                                    \\
\multirow{-4}{*}{\textbf{JE-ZSL} \textit{(ours)}} & \multirow{-4}{*}{MSCOCO}    & \scalebox{0.7}{$d = 2048$} & \cellcolor[HTML]{CBCEFB}\textbf{70.3}                        & \cellcolor[HTML]{CBCEFB}\textbf{60.8}                        & \cellcolor[HTML]{CBCEFB}\textbf{69.1}                        & \multicolumn{1}{c|}{\cellcolor[HTML]{CBCEFB}\textbf{64.7}}                        & 38.4                                                         & \cellcolor[HTML]{CBCEFB}\textbf{28.2}                        & 50.6                                                         & \multicolumn{1}{c|}{\cellcolor[HTML]{CBCEFB}\textbf{36.2}}   \\ \cline{2-11} 
                              &                             & \scalebox{0.7}{$d = 256$}  & \cellcolor[HTML]{CBCEFB}\textbf{68.5}                        & \cellcolor[HTML]{CBCEFB}\textbf{59.0}                        & \cellcolor[HTML]{CBCEFB}\textbf{65.0}                        & \multicolumn{1}{c|}{\cellcolor[HTML]{CBCEFB}\textbf{61.9}}                        & 39.0                                                         & 24.4                                                         & 29.2                                                         & \multicolumn{1}{c|}{26.6}                                    \\
                              &                             & \scalebox{0.7}{$d = 512$}  & \cellcolor[HTML]{CBCEFB}\textbf{70.0}                        & \cellcolor[HTML]{CBCEFB}\textbf{62.1}                        & \cellcolor[HTML]{CBCEFB}\textbf{65.4}                        & \multicolumn{1}{c|}{\cellcolor[HTML]{CBCEFB}\textbf{63.7}}                        & 38.3                                                         & \cellcolor[HTML]{CBCEFB}\textbf{26.4}                        & 41.4                                                         & \multicolumn{1}{c|}{32.2}                                    \\
                              &                             & \scalebox{0.7}{$d = 1024$} & \cellcolor[HTML]{CBCEFB}\textbf{68.4}                        & \cellcolor[HTML]{CBCEFB}\textbf{60.0}                        & \cellcolor[HTML]{CBCEFB}\textbf{65.8}                        & \multicolumn{1}{c|}{\cellcolor[HTML]{CBCEFB}\textbf{62.6}}                        & 40.1                                                         & \cellcolor[HTML]{CBCEFB}\textbf{26.4}                        & 57.3                                                         & \multicolumn{1}{c|}{36.2}                                    \\
\multirow{-4}{*}{\textbf{JE-ZSL} \textit{(ours)}} & \multirow{-4}{*}{ConcCapt}  & \scalebox{0.7}{$d = 2048$} & 67.0                        & \cellcolor[HTML]{CBCEFB}\textbf{61.2}                        & \cellcolor[HTML]{CBCEFB}\textbf{61.5}                        & \multicolumn{1}{c|}{\cellcolor[HTML]{CBCEFB}\textbf{61.3}}                        & \cellcolor[HTML]{CBCEFB}\textbf{41.0}                        & \cellcolor[HTML]{CBCEFB}\textbf{27.4}                        & \cellcolor[HTML]{CBCEFB}\textbf{67.3}                        & \multicolumn{1}{c|}{\cellcolor[HTML]{CBCEFB}\textbf{39.0}}   \\ \hline \hline
\multicolumn{3}{|r|}{ \textit{Baseline} $\colon$ \textit{tf-VAEGAN} \cite{narayan20} }  & \textit{72.2}                                                & \textit{59.8}                                                & \textit{75.1}                                                & \multicolumn{1}{c|}{\textit{66.6}}                                                & \textit{40.8}                                                & \textit{30.8}                                                & \textit{54.6}                                                & \multicolumn{1}{c|}{\textit{39.3}}                           \\ \cline{2-11}
                              &                             & \scalebox{0.7}{$d = 256$}  & 70.0                                                         & 57.8                                                         & 68.2                                                         & \multicolumn{1}{c|}{62.6}                                                         & 36.4                                                         & 25.5                                                         & 42.2                                                         & \multicolumn{1}{c|}{31.8}                                    \\
                              &                             & \scalebox{0.7}{$d = 512$}  & 67.4                                                         & 57.3                                                         & 73.8                                                         & \multicolumn{1}{c|}{64.5}                                                         & 38.9                                                         & 26.8                                                         & 50.9                                                         & \multicolumn{1}{c|}{35.0}                                    \\
                              &                             & \scalebox{0.7}{$d = 1024$} & 69.9                                                         & \cellcolor[HTML]{CBCEFB}\textbf{63.1}                        & 71.5                                                         & \multicolumn{1}{c|}{\cellcolor[HTML]{CBCEFB}\textbf{67.0}}                        & \cellcolor[HTML]{CBCEFB}\textbf{41.4}                        & 29.2                                                         & 53.0                                                         & \multicolumn{1}{c|}{37.6}                                    \\
\multirow{-4}{*}{\textbf{JE-ZSL} \textit{(ours)}} & \multirow{-4}{*}{Flickr30K} & \scalebox{0.7}{$d = 2048$} & 69.9                                                         & \cellcolor[HTML]{CBCEFB}\textbf{63.1}                        & 61.5                                                         & \multicolumn{1}{c|}{\cellcolor[HTML]{CBCEFB}\textbf{67.0}}                        & 40.2                                                         & 29.9                                                         & 47.1                                                         & \multicolumn{1}{c|}{36.8}                                    \\ \cline{2-11} 
                              &                             & \scalebox{0.7}{$d = 256$}  & 70.5                                                         & 56.9                                                         & 73.4                                                         & \multicolumn{1}{c|}{64.0}                                                         & \cellcolor[HTML]{CBCEFB}\textbf{43.2}                        & \cellcolor[HTML]{CBCEFB}\textbf{30.0}                        & \cellcolor[HTML]{CBCEFB}\textbf{62.3}                        & \multicolumn{1}{c|}{\cellcolor[HTML]{CBCEFB}\textbf{40.4}}   \\
                              &                             & \scalebox{0.7}{$d = 512$}  & 72.3                                                         & \cellcolor[HTML]{CBCEFB}\textbf{60.6}                        & 64.0                                                         & \multicolumn{1}{c|}{66.0}                                                         & \cellcolor[HTML]{CBCEFB}\textbf{43.7}                        & 30.6                                                         & \cellcolor[HTML]{CBCEFB}\textbf{59.7}                        & \multicolumn{1}{c|}{\cellcolor[HTML]{CBCEFB}\textbf{40.5}}   \\
                              &                             & \scalebox{0.7}{$d = 1024$} & 71.8                                                         & \cellcolor[HTML]{CBCEFB}\textbf{62.8}                        & 63.8                                                         & \multicolumn{1}{c|}{\cellcolor[HTML]{CBCEFB}\textbf{66.8}}                        & 40.8                                                         & 30.8                                                         & \cellcolor[HTML]{CBCEFB}\textbf{53.6}                        & \multicolumn{1}{c|}{\cellcolor[HTML]{CBCEFB}\textbf{38.5}}   \\
\multirow{-4}{*}{\textbf{JE-ZSL} \textit{(ours)}} & \multirow{-4}{*}{MSCOCO}    & \scalebox{0.7}{$d = 2048$} & \cellcolor[HTML]{CBCEFB}\textbf{76.3}                        & \cellcolor[HTML]{CBCEFB}\textbf{62.3}                        & \cellcolor[HTML]{CBCEFB}\textbf{78.3}                        & \multicolumn{1}{c|}{\cellcolor[HTML]{CBCEFB}\textbf{69.4}}                        & \cellcolor[HTML]{CBCEFB}\textbf{41.8}                        & 27.3                                                         & \cellcolor[HTML]{CBCEFB}\textbf{70.6}                        & \multicolumn{1}{c|}{\cellcolor[HTML]{CBCEFB}\textbf{39.4}}   \\ \cline{2-11} 
                              &                             & \scalebox{0.7}{$d = 256$}  & 71.9                                                         & \cellcolor[HTML]{CBCEFB}\textbf{61.0}                        & 73.6                                                         & \multicolumn{1}{c|}{\cellcolor[HTML]{CBCEFB}\textbf{66.8}}                        & \cellcolor[HTML]{CBCEFB}\textbf{41.8}                        & 29.3                                                         & 51.9                                                         & \multicolumn{1}{c|}{37.5}                                    \\
                              &                             & \scalebox{0.7}{$d = 512$}  & 69.4                                                         & 58.8                                                         & 71.2                                                         & \multicolumn{1}{c|}{64.4}                                                         & 39.6                                                         & 29.8                                                         & 52.5                                                         & \multicolumn{1}{c|}{38.0}                                    \\
                              &                             & \scalebox{0.7}{$d = 1024$} & 68.1                                                         & 59.0                                                         & 71.8                                                         & \multicolumn{1}{c|}{64.8}                                                         & 40.3                                                         & \cellcolor[HTML]{CBCEFB}\textbf{31.9}                        & \cellcolor[HTML]{CBCEFB}\textbf{54.7}                        & \multicolumn{1}{c|}{\cellcolor[HTML]{CBCEFB}\textbf{40.3}}   \\
\multirow{-4}{*}{\textbf{JE-ZSL} \textit{(ours)}} & \multirow{-4}{*}{ConcCapt}  & \scalebox{0.7}{$d = 2048$} & 70.6                                                         & \cellcolor[HTML]{CBCEFB}\textbf{63.0}                        & \cellcolor[HTML]{CBCEFB}\textbf{75.6}                        & \multicolumn{1}{c|}{\cellcolor[HTML]{CBCEFB}\textbf{68.7}}                        & \cellcolor[HTML]{CBCEFB}\textbf{42.1}                        & \cellcolor[HTML]{CBCEFB}\textbf{32.1}                        & 49.5                                                         & \multicolumn{1}{c|}{39.0}                                    \\ \hline
\end{tabular}
\caption{Joint Embeddings for zero-shot learning (\textbf{JE-ZSL}) - ablation study on the effect of the dimensionality. We mark in blue and bold any improvement scored by joint embeddings over the respective backbone zero-shot learning method used as baseline.}
\label{tab:abl1}
\end{table*}

\subsection{Effect of the Embedding Size}

In Table \ref{tab:abl1}, we evaluate the impact of various embedding sizes applied to JE-ZSL. We take advantage of state-of-the-art methods (CDL \cite{Jiang18}, f-CLSWGAN \cite{xian2018feature} and tf-VAEGAN \cite{narayan20}) on coarse-grained benchmark datasets: AWA$2$~\cite{AWA} and aPY \cite{APY}. These methods are fed with off-the-shelves conventional visual features, exploiting the standard ResNet$-101$ embeddings as provided by \cite{xian2018feature}. We then compare the very same zero-shot learning approaches as a baseline references to evaluate the impact in performance of having visual embeddings trained in a joint manner with text information, taking advantage of the architecture presented in Section \ref{sez:arch}. For the joint training, we take advantage of Flickr30K~\cite{plummer2015flickr30k}, MSCOCO~\cite{lin2014microsoft} and ConcCapt~\cite{sharma2018conceptual}  and, in all cases, we extract joint embedding of dimensions $d=256,512,1024$ and $2048$ to compare them with baseline visual embeddings. \\

\begin{figure*}[t!]
    \centering
    \includegraphics [width=\textwidth]{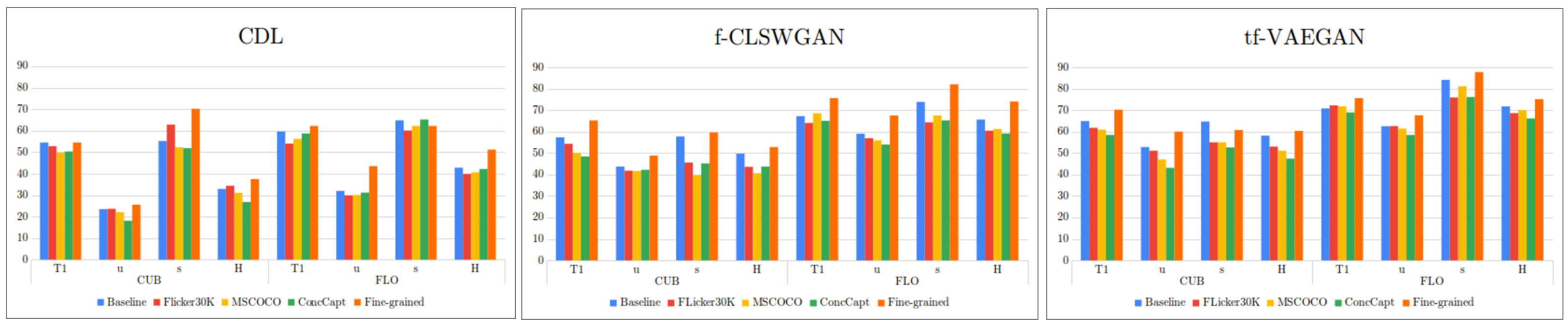}
    \caption{Joint embeddings for zero-shot learning (\textbf{JE-ZSL}) to handle fine-grained datasets.
    }
    \label{fig:fine-grained-results}
\end{figure*}
\textbf{Discussion}. Table~\ref{tab:abl1} shows the results in bold text font and color which are superior to the respective baseline values:
It is evident that the joint training produces a widespread improvement on performance, being consistent with respect to different baseline approaches and multimodal datasets used. Notably, the idea of performing joint training for the embedding has the peculiar trait of ensuring a stable performance even at low embedding sizes. No matter which is the input embedding size, in fact, in a few  cases, joint training with $d=256$, $d=512$, $d=1024$ and $d=2048$ is capable of improving upon the baseline methods in both standard zero-shot learning (T$1$ on aPY for both Flickr30K and MSCOCO with respect to CDL zero-shot learn method, T$1$ on AWA$2$ for MSCOCO with respect to f-CLSWGAN) and generalized zero-shot learning as well ($H$ for MSCOCO and ConcCapt with respect to f-CLSWGAN and $H$ for MSCOCO with respect to tf-VAEGAN). This is especially remarkable if considering that baseline approaches exploit $2048-$dimensional visual embeddings: Even if using lower dimensional embeddings, there are only very few cases in which severe drops are registered (such as for $d=256$ on Flickr30K with respect to f-CLSWGAN) and performance are often better. Overall, methods such as CDL are improved in their standard and generalized zero-shot recognition by $+2.9\%$ (T1) and $+5\%$ ($H$), respectively using ConcCapt and $d=2048$. Similarly, using the very same setup, we register analogous improvements of $+0.5\%$ for T$1$ and $+3.0\%$ for $H$ (with respect to f-CLSWAGAN), while we enhance tf-VAEGAN by $+4.1\%$ (T1) on standard zero-shot learning and $+2.8\%$ ($H$) for generalized zero-shot learning using $d=2048$ and MSCOCO. These results indicate that the joint training improve representation with visual and semantic cues which in turn eases zero-shot learning recognition task.

\begin{table*}[t!]
\resizebox{\textwidth}{!}{%
\centering
\begin{tabular}{|ll|cccc|cccc|cccc|cccc|} \hline
                                         &             & \multicolumn{4}{c}{AWA$2$~\cite{AWA}}   & \multicolumn{4}{|c|}{CUB~\cite{CUB}}   & \multicolumn{4}{|c|}{FLO~\cite{FLO}}   & \multicolumn{4}{|c|}{aPY~\cite{APY}}   \\
                                         &             & T1   & $u$    & $s$    & $H$    & T1   & $u$    & $s$    & $H$    & T1   & $u$    & $s$    & $H$    & T1   & $u$ & $s$ & $H$    \\ \hline \hline
\multirow{7}{*}{\rotatebox{90}{Compatibility Func}} & LATEM \cite{XianCVPR16}       & 55.1 & 7.3  & 71.7 & 13.3 & 49.6 & 15.2 & 57.3 & 24   & 40.4 & 6.6  & 47.6 & 11.5 & 36.8 & 5.7  & 65.6 & 10.4 \\
                                         & SynC \cite{Changpinyo16}       & 49.3 & 8.9  & \cellcolor[HTML]{CBCEFB}\textbf{87.3} & 16.2 & 53.0 & 11.5 & \cellcolor[HTML]{CBCEFB}\textbf{70.9} & 19.8 & $-$  & $-$  & $-$  & $-$  & 23.9 & 7.4  & 66.3 & 13.3 \\
                                         & SJE \cite{Akata15}         & 65.6 & 11.3 & 74.6 & 19.6 & 53.9 & 23.5 & 59.2 & 33.6 & 53.4 & 13.9 & 47.6 & 21.5 & 31.7 & 1.3  & 71.4 & 2.6  \\
                                         & DeViSE \cite{Frome13}      & 54.2 & 13.4 & 68.7 & 22.4 & 52.0 & 23.8 & 53.0 & 32.8 & 45.9 & 9.9  & 44.2 & 16.2 & 37.0 & 3.5  & \cellcolor[HTML]{CBCEFB}\textbf{78.4} & 6.7  \\
                                         & KerZSL \cite{Zhang18}      & 70.5 & 18.9 & 82.7 & 30.8 & 51.7 & 21.6 & 52.8 & 30.6 & $-$  & $-$  & $-$  & $-$  & 45.3 & 10.5 & 76.2 & 18.5 \\
                                         & CDL \cite{Jiang18}         & 69.9 & 28.1 & 73.5 & 40.6 & \cellcolor[HTML]{CBCEFB}\textbf{54.5} & 23.5 & 55.2 & 32.9 & 59.6 & 32.0 & 64.8 & 42.8 & 43.0 & 19.8 & 48.6 & 28.1 \\
                                         \cline{2-18}
                                         & \textbf{JE-ZSL} \textit{(ours)}   & \cellcolor[HTML]{CBCEFB}\textbf{71.0} & \cellcolor[HTML]{CBCEFB}\textbf{29.3} & 77.1 & \cellcolor[HTML]{CBCEFB}\textbf{42.5} & 54.1 & \cellcolor[HTML]{CBCEFB}\textbf{25.6} & 70.2 & \cellcolor[HTML]{CBCEFB}\textbf{37.5} & \cellcolor[HTML]{CBCEFB}\textbf{62.2} & \cellcolor[HTML]{CBCEFB}\textbf{43.5} & \cellcolor[HTML]{CBCEFB}\textbf{62.2} & \cellcolor[HTML]{CBCEFB}\textbf{51.2} & \cellcolor[HTML]{CBCEFB}\textbf{45.9} & \cellcolor[HTML]{CBCEFB}\textbf{24.5} & 51.1 & \cellcolor[HTML]{CBCEFB}\textbf{33.1} \\ \hline \hline
\multirow{10}{*}{\rotatebox{90}{Feature Generation}}     & f-CLSWGAN \cite{xian2018feature}  & 68.2 & 57.9 & 61.4 & 59.6 & 57.3 & 43.7 & 57.7 & 49.7 & 67.2 & 59.0 & 73.8 & 65.6 & 40.5 & 25.8 & 59.5 & 36.0 \\
                                         & Cycle-WGAN \cite{felix2018multi}  & 66.8 & \cellcolor[HTML]{CBCEFB}\textbf{63.4} & 59.6 & 59.8 & 58.6 & 59.3 & 47.9 & 53.0 & 70.3 & 61.6 & 69.2 & 65.2 & $-$  & $-$  & $-$  & $-$  \\
                                         & f-VAEGAN-D2 \cite{Xian_2019_CVPR} & 71.1 & 57.6 & 70.6 & 63.5 & 61.0 & 48.4 & 60.1 & 53.6 & 67.7 & 56.8 & 74.9 & 64.6 & $-$  & $-$  & $-$  & $-$  \\
                                         & CADA-VAE \cite{Schonfeld19} & 68.2 & 55.8 & 75.0 & 63.9 & 57.3 & 51.6 & 53.5 & 52.4 & $-$  & $-$  & $-$  & $-$  & $-$  & $-$  & $-$  & $-$  \\
                                         & GDAN \cite{Huang19}        & 68.1 & 32.1 & 67.5 & 43.5 & 58.8 & 39.3 & \cellcolor[HTML]{CBCEFB}\textbf{66.7} & 49.5 & $-$  & $-$  & $-$  & $-$  & 38.3 & 30.4 & \cellcolor[HTML]{CBCEFB}\textbf{75.0} & \cellcolor[HTML]{CBCEFB}\textbf{43.4} \\
                                         & COSMO \cite{Atzmon19} & $-$  & 52.8 & \cellcolor[HTML]{CBCEFB}\textbf{80.0} & 63.6 & $-$  & 44.4 & 57.8 & 50.2 & $-$  & 59.6 & 81.4 & 68.8 & $-$  & $-$  & $-$  & $-$  \\
                                         & PGN \cite{Yu20}         & 71.2 & 48.0 & 83.6 & 61.0 & 68.3 & 48.5 & 57.2 & 52.5 & \cellcolor[HTML]{CBCEFB}\textbf{81.4} & 63.6 & 77.8 & 70.0 & $-$  & $-$  & $-$  & $-$  \\
                                         & 0-VAEGAN \cite{Han20}   & 69.9 & 56.2 & 71.7 & 63.0 & 54.9 & 41.1 & 48.5 & 44.4 & $-$  & $-$  & $-$  & $-$  & 36.3 & 31.7 & 53.2 & 39.7 \\
                                         & tf-VAEGAN \cite{narayan20}   & \cellcolor[HTML]{CBCEFB}\textbf{72.2} & 59.8 & 75.1 & 66.6 & 64.9 & 52.8 & 64.7 & 58.1 & 70.8 & 62.5 & 84.1 & 71.7 & 40.8 & 30.8 & 54.6 & 39.3 \\ \cline{2-18}
                                         & \textbf{JE-ZSL} \textit{(ours)}   & 70.6 & 63.0 & 75.6 & \cellcolor[HTML]{CBCEFB}\textbf{68.7} & \cellcolor[HTML]{CBCEFB}\textbf{70.2} & \cellcolor[HTML]{CBCEFB}\textbf{60.0} & 60.7 & \cellcolor[HTML]{CBCEFB}\textbf{60.3} & 75.6 & \cellcolor[HTML]{CBCEFB}\textbf{67.6} & \cellcolor[HTML]{CBCEFB}\textbf{84.7} & \cellcolor[HTML]{CBCEFB}\textbf{75.1} & \cellcolor[HTML]{CBCEFB}\textbf{42.1} & \cellcolor[HTML]{CBCEFB}\textbf{32.1} & 49.5 & 39.0 \\ \hline
\end{tabular}}
\caption{Comparison of our proposed \textbf{JE-ZSL} against state-of-the-art methods based on either compatibility functions or feature generation. Improvements scored by \textbf{JE-ZSL} over each of this class of methods is highlighted in blue and bolded.}
\label{tab:soa}
\end{table*}

\subsection{Fine-grained Joint Embeddings}\label{sez:fine-grained}
Fig.~\ref{fig:fine-grained-results} shows the comparison of the setup $d=2048$ with Flickr30K~\cite{plummer2015flickr30k}, MSCOCO~\cite{lin2014microsoft},  ConcCapt~\cite{sharma2018conceptual} and fine-grained datasets (CUB \cite{CUB} and FLO \cite{FLO}) for which captions annotating the images are available on seen classes. This helps us in understanding to which extent a joint training is capable of enhancing results, while disentangling it from the intrinsic difficulty of the dataset on which the ZSL evaluation is carried out. 
In fact, on CUB and FLO datasets, categories are fine-grained in nature (bird and flower species, respectively) and, therefore, much harder to classify - and we will evaluate joint embeddings also in this playground.\\
\textbf{Discussion}. Despite the fine-grained nature of the classes to recognize, dataset such as Flickr30K are enough to improve $s$ and $H$ metrics with respect to the baseline model CDL (on CUB). Joint embeddings trained using ConcCapt improve the metric $s$ scored by CDL on FLO. Using MSCOCO, joint embeddings boost the performance of f-CLSWGAN (T1 on FLO) and tf-VAEGAN is improved as well (T1 on FLO) using either Flickr30K or MSCOCO. Thus, despite the challenging nature of fine-grained recognition task, still the idea of joint training provides subtle improvements.

These preliminary improvements are furthermore enhanced as soon as the fine-grained textual information is exploited on CUB and FLO: We improve CDL, f-CLSWGAN and tf-VAEGAN. More specifically, our proposed approach improves the T1 score of f-CLSWGAN on CUB ($+7.9$). On FLO dataset, CDL, f-CLSWGAN and tf-VAEGAN baseline methods are improved by $+2.6\%$, $+8.4\%$ and $+4.8\%$ in their zero-shot learning performance (T1). Similar improvements are registered in GZSL: the $H$ value by CDL is improved by $+4.6\%$ on CUB and by $+8.4\%$ on FLO. The $H$ value by f-CLSWGAN is improved by $+3.1\%$ on CUB and $+8.4\%$ on FLO. Finally, tf-VAEGAN achieves an improvement of $H$ on both CUB ($+2.2\%$) and FLO ($+3.4\%$). The scored performance improvements on CUB and FLO datasets demonstrate that fine-grained textual information enhances visual embeddings when paired with our proposed joint training.

\section{Comparison with the State-of-the-Art}

In this section, we test our proposed joint embeddings against state-of-the-art zero-shot learning methods based on two mainstream paradigms: Compatibility functions and feature generation.\\
\textbf{\em Compatibility functions.} Among the state-of-the-art approaches, we included the idea of generating a latent representation on which visual and semantic information is projected (LATEM \cite{XianCVPR16}) and the generation of synthetic classifiers for the unseen classes as a linear combination of classifiers of the seen ones (SynC \cite{Changpinyo16}). We also compare against structured embedding obtained by combining different sources of side information (SJE \cite{Akata15}) and the deep visual and semantic embedding model known as DeVISE \cite{Frome13}.  We include the kernelized approach to learn a compatibility function in  max-margin sense (KerZSL \cite{Zhang18}) and the coupled dictionary learning approach to learn and then match a set of visual and semantic atoms with which images and side information can be described (CDL \cite{Jiang18}). \\
\textbf{\em Feature generating approaches.} We consider several methods to generate visual features for the unseen classes, known in terms of attributes only. Namely, the Wasserstein GAN trained with a classifier loss (f-CLSWGAN \cite{Xian_2018_CVPR}) and its variant in which a cycle-consistency loss is added to predict attributes from synthesized features. (Cycle-WGAN \cite{Felix18}). We compare against the usage of confidence smoothing (COSMO \cite{Atzmon19}) which, combined with a feature generation approach, helps in balancing the recognition performance across seen and unseen classes. A similar effect is achieved by the adoption of dual learning to train feature generation and attribute prediction synchronously (GDAN \cite{Huang19}). We also compared against the idea of scheduling the generation of synthetic feature by means of episodic training (PGN \cite{Yu20}). In addition, we consider methods in which the feature generation stage is enhanced by the adoption of a variational autoencoder to pair a GAN: 0-VAE-GAN \cite{Gao20}, f-VAEGAN-D2 \cite{Xian19}, CADA-VAE \cite{Schonfeld19}, and tf-VAEGAN \cite{narayan20}.

Table~\ref{tab:soa} shows the results of this analysis.
We fix the dimensionality of the embedding to $d=2048$ that were used to boost CDL when challenging other compatibility functions.
When comparing against tf-VAEGAN, performance is enhanced with  joint embedding while comparing against other feature generating approaches. The fine-grained setup (see Section \ref{sez:fine-grained}) was used when handling both CUB and FLO datasets, while using Conceptual Captions for the joint training on AWA$2$ and aPY.\\
\textbf{Discussion}. With our proposed semantically grounded embeddings, the performance previously scored by compatibility functions is considerably improved. 
We observed performance boosts with respect to the metric \textit{u}  over all datasets (AWA$2$, CUB, FLO and aPY) along with \textit{s} on FLO dataset.
Additionally, on AWA$2$, the best performance on zero-shot learning (KerZSL) is improved by $+0.5\%$ and the $H$ of the best scoring method on generalized zero-shot learning (CDL) is boosted by $+1.9\%$. Similarly, the $H$ metric is enhanced for generalized zero-shot learning on CUB ($+5.9\%$ with respect to SJE) on FLO ($+8.4\%$ with respect to CDL) and on aPY ($+5\%$ with respect to CDL). The best scored T1 classification accuracy of prior methods is improved on FLO ($+2.6\%$ with respect to CDL) and on aPY ($+0.6\%$ with respect to KerZSL). While considering feature generating approaches, our proposed embedding boosts the $H$ score on both AWA$2$ and CUB ($+2.1\%$ and $+2.2\%$, respectively - both with respect to tf-VAEGAN). On aPY, we also improve f-CLSWGAN in the T1 score for ZSL ($+1.6\%$).

\section{Conclusions}
This paper presents joint embeddings for zero-shot learning tasks. Unlike existing methods which utilize \textit{fixed} visual and semantic information, we proposed to learn semantically grounded visual embedding by capturing interactions between images and text clues with a two-stream network on a proxy task leveraging large-scale unlabelled data sources. We evaluated our embeddings with various state-of-the-art methods on two fine-grained (CUB and FLO) and two coarse-grained (AWA$2$ and aPY) benchmark zero-shot learning datasets. Our evaluation showed that our proposed embeddings considerably improved standard and generalised zero-shot learning performance across various compatibility functions and GAN-based methods.

{\small
\bibliographystyle{ieee_fullname}
\bibliography{egbib}

\begin{thebibliography}{10}\itemsep=-1pt

\bibitem{Jiang:ICCV19}
Transferable contrastive network for generalized zero-shot learning.
\newblock In {\em The IEEE International Conference on Computer Vision (ICCV)},
  2019.

\bibitem{Akata15}
Zeynep Akata, Scott Reed, Daniel Walter, Honglak Lee, and Bernt Schiele.
\newblock Evaluation of output embeddings for fine-grained image
  classification.
\newblock In {\em IEEE Conference on Computer Vision and Pattern Recognition},
  2015.

\bibitem{antol2015vqa}
Stanislaw Antol, Aishwarya Agrawal, Jiasen Lu, Margaret Mitchell, Dhruv Batra,
  C Lawrence~Zitnick, and Devi Parikh.
\newblock Vqa: Visual question answering.
\newblock In {\em Proceedings of the IEEE international conference on computer
  vision}, pages 2425--2433, 2015.

\bibitem{arshad2019aiding}
Omer Arshad, Ignazio Gallo, Shah Nawaz, and Alessandro Calefati.
\newblock Aiding intra-text representations with visual context for multimodal
  named entity recognition.
\newblock In {\em 2019 15th IAPR International Conference on Document Analysis
  and Recognition (ICDAR)}. IEEE, 2019.

\bibitem{Atzmon19}
Yuval Atzmon and Gal Chechik.
\newblock Adaptive confidence smoothing for generalized zero-shot learning.
\newblock In {\em IEEE Conference on Computer Vision and Pattern Recognition
  (CVPR)}, 2019.

\bibitem{baltrusaitis2019multimodal}
Tadas Baltrusaitis, Chaitanya Ahuja, and Louis-Philippe Morency.
\newblock Multimodal machine learning: A survey and taxonomy.
\newblock {\em IEEE transactions on pattern analysis and machine intelligence},
  41(2):423--443, 2019.

\bibitem{cacheux2019modeling}
Yannick~Le Cacheux, Herve~Le Borgne, and Michel Crucianu.
\newblock Modeling inter and intra-class relations in the triplet loss for
  zero-shot learning.
\newblock In {\em Proceedings of the IEEE International Conference on Computer
  Vision}, pages 10333--10342, 2019.

\bibitem{Changpinyo16}
Soravit Changpinyo, Wei-Lun Chao, Boqing Gong, and Fei Sha.
\newblock Synthesized classifiers for zero-shot learning.
\newblock In {\em IEEE Conference on Computer Vision and Pattern Recognition
  (CVPR)}, 2016.

\bibitem{changpinyo2018classifier}
Soravit Changpinyo, Wei-Lun Chao, Boqing Gong, and Fei Sha.
\newblock Classifier and exemplar synthesis for zero-shot learning.
\newblock {\em The Springer International Journal of Computer Vision (IJCV)},
  2019.

\bibitem{chen2020uniter}
Yen-Chun Chen, Linjie Li, Licheng Yu, Ahmed El~Kholy, Faisal Ahmed, Zhe Gan, Yu
  Cheng, and Jingjing Liu.
\newblock Uniter: Universal image-text representation learning.
\newblock ECCV, 2020.

\bibitem{devlin2018bert}
Jacob Devlin, Ming-Wei Chang, Kenton Lee, and Kristina Toutanova.
\newblock Bert: Pre-training of deep bidirectional transformers for language
  understanding.
\newblock {\em arXiv preprint arXiv:1810.04805}, 2018.

\bibitem{faghri2017vse++}
Fartash Faghri, David~J Fleet, Jamie~Ryan Kiros, and Sanja Fidler.
\newblock Vse++: Improving visual-semantic embeddings with hard negatives.
\newblock {\em arXiv preprint arXiv:1707.05612}, 2017.

\bibitem{farhadi2009describing}
Ali Farhadi, Ian Endres, Derek Hoiem, and David Forsyth.
\newblock Describing objects by their attributes.
\newblock In {\em 2009 IEEE Conference on Computer Vision and Pattern
  Recognition}, pages 1778--1785. IEEE, 2009.

\bibitem{felix2018multi}
Rafael Felix, Vijay~BG Kumar, Ian Reid, and Gustavo Carneiro.
\newblock Multi-modal cycle-consistent generalized zero-shot learning.
\newblock In {\em The European Conference on Computer Vision (ECCV)}, 2018.

\bibitem{Felix18}
Rafael Felix, Vijay~BG Kumar, Ian Reid, and Gustavo Carneiro.
\newblock Multi-modal cycle-consistent generalized zero-shot learning.
\newblock In {\em European Conference on Computer Vision (ECCV)}, 2018.

\bibitem{Frome13}
Andrea Frome, Greg~S Corrado, Jon Shlens, Samy Bengio, Jeff Dean, Marc'Aurelio
  Ranzato, and Tomas Mikolov.
\newblock {DeVISE}: A deep visual-semantic embedding model.
\newblock In {\em Advances in Neural Information Processing Systems (NeurIPS)},
  2013.

\bibitem{gallo2017multimodal}
Ignazio Gallo, Alessandro Calefati, and Shah Nawaz.
\newblock Multimodal classification fusion in real-world scenarios.
\newblock In {\em 2017 14th IAPR International Conference on Document Analysis
  and Recognition (ICDAR)}, volume~5, pages 36--41. IEEE, 2017.

\bibitem{Gao20}
Rui Gao, Xingsong Hou, Jie Qin, Jiaxin Chen, Li Liu, Fan Zhu, Zhao Zhang, and
  Ling Shao.
\newblock Zero-vae-gan: Generating unseen features for generalized and
  transductive zero-shot learning.
\newblock {\em IEEE Transactions on Image Processing}, 29:3665--3680, 2020.

\bibitem{guo2019dual}
Yuchen Guo, Guiguang Ding, Jungong Han, Xiaohan Ding, Sicheng Zhao, Zheng Wang,
  Chenggang Yan, and Qionghai Dai.
\newblock Dual-view ranking with hardness assessment for zero-shot learning.
\newblock In {\em The AAAI Conference on Artificial Intelligence}, volume~33,
  pages 8360--8367, 2019.

\bibitem{Han20}
Zongyan Han, Zhenyong Fu, and Jian Yang.
\newblock Learning the redundancy-free features for generalized zero-shot
  object recognition.
\newblock In {\em IEEE Conference on Computer Vision and Pattern Recognition
  (CVPR)}, 2020.

\bibitem{he2016deep}
Kaiming He, Xiangyu Zhang, Shaoqing Ren, and Jian Sun.
\newblock Deep residual learning for image recognition.
\newblock In {\em Proceedings of the IEEE conference on computer vision and
  pattern recognition}, pages 770--778, 2016.

\bibitem{huang2019generative}
He Huang, Changhu Wang, Philip~S Yu, and Chang-Dong Wang.
\newblock Generative dual adversarial network for generalized zero-shot
  learning.
\newblock In {\em The IEEE Conference on Computer Vision and Pattern
  Recognition (CVPR)}, 2019.

\bibitem{Huang19}
He Huang, Changhu Wang, Philip~S Yu, and Chang-Dong Wang.
\newblock Generative dual adversarial network for generalized zero-shot
  learning.
\newblock In {\em IEEE Conference on Computer Vision and Pattern Recognition
  (CVPR)}, 2019.

\bibitem{jiang2018learning}
Huajie Jiang, Ruiping Wang, Shiguang Shan, and Xilin Chen.
\newblock Learning class prototypes via structure alignment for zero-shot
  recognition.
\newblock In {\em European conference on computer vision (ECCV)}, 2018.

\bibitem{Jiang18}
Huajie Jiang, Ruiping Wang, Shiguang Shan, and Xilin Chen.
\newblock Learning class prototypes via structure alignment for zero-shot
  recognition.
\newblock In {\em European conference on computer vision (ECCV)}, 2018.

\bibitem{kodirov2017semantic}
Elyor Kodirov, Tao Xiang, and Shaogang Gong.
\newblock Semantic autoencoder for zero-shot learning.
\newblock In {\em The IEEE Conference on Computer Vision and Pattern
  Recognition (CVPR)}, 2017.

\bibitem{krishna2017visual}
Ranjay Krishna, Yuke Zhu, Oliver Groth, Justin Johnson, Kenji Hata, Joshua
  Kravitz, Stephanie Chen, Yannis Kalantidis, Li-Jia Li, David~A Shamma, et~al.
\newblock Visual genome: Connecting language and vision using crowdsourced
  dense image annotations.
\newblock {\em International journal of computer vision}, 123(1):32--73, 2017.

\bibitem{AWA}
CH. Lampert, H. Nickisch, and S. Harmeling.
\newblock Learning to detect unseen object classes by between-class attribute
  transfer.
\newblock In {\em The IEEE Conference on Computer Vision and Pattern
  Recognition (CVPR)}, 2009.

\bibitem{li2019visualbert}
Liunian~Harold Li, Mark Yatskar, Da Yin, Cho-Jui Hsieh, and Kai-Wei Chang.
\newblock Visualbert: A simple and performant baseline for vision and language.
\newblock {\em arXiv preprint arXiv:1908.03557}, 2019.

\bibitem{li2018discriminative}
Yan Li, Junge Zhang, Jianguo Zhang, and Kaiqi Huang.
\newblock Discriminative learning of latent features for zero-shot recognition.
\newblock In {\em The IEEE Conference on Computer Vision and Pattern
  Recognition (CVPR)}, 2018.

\bibitem{lin2014microsoft}
Tsung-Yi Lin, Michael Maire, Serge Belongie, James Hays, Pietro Perona, Deva
  Ramanan, Piotr Doll{\'a}r, and C~Lawrence Zitnick.
\newblock Microsoft coco: Common objects in context.
\newblock In {\em European conference on computer vision}, pages 740--755.
  Springer, 2014.

\bibitem{lu2019vilbert}
Jiasen Lu, Dhruv Batra, Devi Parikh, and Stefan Lee.
\newblock Vilbert: Pretraining task-agnostic visiolinguistic representations
  for vision-and-language tasks.
\newblock In {\em Advances in Neural Information Processing Systems}, pages
  13--23, 2019.

\bibitem{mikolov2013distributed}
Tomas Mikolov, Ilya Sutskever, Kai Chen, Greg~S Corrado, and Jeff Dean.
\newblock Distributed representations of words and phrases and their
  compositionality.
\newblock In {\em Advances in neural information processing systems}, pages
  3111--3119, 2013.

\bibitem{narayan2020latent}
Sanath Narayan, Akshita Gupta, Fahad~Shahbaz Khan, Cees~GM Snoek, and Ling
  Shao.
\newblock Latent embedding feedback and discriminative features for zero-shot
  classification.
\newblock {\em arXiv preprint arXiv:2003.07833}, 2020.

\bibitem{narayan20}
Sanath Narayan, Akshita Gupta, Fahad~Shahbaz Khan, Cees~GM Snoek, and Ling
  Shao.
\newblock Latent embedding feedback and discriminative features for zero-shot
  classification.
\newblock In {\em The European Conference on Computer Vision (ECCV)}, 2020.

\bibitem{nawaz2019cross}
Shah Nawaz, Muhammad Kamran~Janjua, Ignazio Gallo, Arif Mahmood, Alessandro
  Calefati, and Faisal Shafait.
\newblock Do cross modal systems leverage semantic relationships?
\newblock In {\em Proceedings of the IEEE International Conference on Computer
  Vision Workshops}, pages 0--0, 2019.

\bibitem{FLO}
M-E. Nilsback and A. Zisserman.
\newblock A visual vocabulary for flower classification.
\newblock In {\em The IEEE Conference on Computer Vision and Pattern
  Recognition (CVPR)}, 2006.

\bibitem{nilsback2008automated}
Maria-Elena Nilsback and Andrew Zisserman.
\newblock Automated flower classification over a large number of classes.
\newblock In {\em 2008 Sixth Indian Conference on Computer Vision, Graphics \&
  Image Processing}, pages 722--729. IEEE, 2008.

\bibitem{plummer2015flickr30k}
Bryan~A Plummer, Liwei Wang, Chris~M Cervantes, Juan~C Caicedo, Julia
  Hockenmaier, and Svetlana Lazebnik.
\newblock Flickr30k entities: Collecting region-to-phrase correspondences for
  richer image-to-sentence models.
\newblock In {\em Proceedings of the IEEE international conference on computer
  vision}, pages 2641--2649, 2015.

\bibitem{qi2020imagebert}
Di Qi, Lin Su, Jia Song, Edward Cui, Taroon Bharti, and Arun Sacheti.
\newblock Imagebert: Cross-modal pre-training with large-scale weak-supervised
  image-text data.
\newblock {\em arXiv preprint arXiv:2001.07966}, 2020.

\bibitem{reed2016learning}
Scott Reed, Zeynep Akata, Honglak Lee, and Bernt Schiele.
\newblock Learning deep representations of fine-grained visual descriptions.
\newblock In {\em Proceedings of the IEEE Conference on Computer Vision and
  Pattern Recognition}, pages 49--58, 2016.

\bibitem{ESZSL}
Bernardino Romera-Paredes and Philip Torr.
\newblock An embarrassingly simple approach to zero-shot learning.
\newblock In {\em The International Conference on Machine Learning (ICML)},
  2015.

\bibitem{Schonfeld_2019_CVPR}
Edgar Schonfeld, Sayna Ebrahimi, Samarth Sinha, Trevor Darrell, and Zeynep
  Akata.
\newblock Generalized zero- and few-shot learning via aligned variational
  autoencoders.
\newblock In {\em The IEEE Conference on Computer Vision and Pattern
  Recognition (CVPR)}, June 2019.

\bibitem{Schonfeld19}
Edgar Schonfeld, Sayna Ebrahimi, Samarth Sinha, Trevor Darrell, and Zeynep
  Akata.
\newblock Generalized zero- and few-shot learning via aligned variational
  autoencoders.
\newblock In {\em IEEE Conference on Computer Vision and Pattern Recognition
  (CVPR)}, 2019.

\bibitem{sharma2018conceptual}
Piyush Sharma, Nan Ding, Sebastian Goodman, and Radu Soricut.
\newblock Conceptual captions: A cleaned, hypernymed, image alt-text dataset
  for automatic image captioning.
\newblock In {\em Proceedings of the 56th Annual Meeting of the Association for
  Computational Linguistics (Volume 1: Long Papers)}, pages 2556--2565, 2018.

\bibitem{simonyan2014very}
Karen Simonyan and Andrew Zisserman.
\newblock Very deep convolutional networks for large-scale image recognition.
\newblock {\em arXiv preprint arXiv:1409.1556}, 2014.

\bibitem{song2018transductive}
Jie Song, Chengchao Shen, Yezhou Yang, Yang Liu, and Mingli Song.
\newblock Transductive unbiased embedding for zero-shot learning.
\newblock In {\em The IEEE Conference on Computer Vision and Pattern
  Recognition (CVPR)}, 2018.

\bibitem{tan2019lxmert}
Hao Tan and Mohit Bansal.
\newblock Lxmert: Learning cross-modality encoder representations from
  transformers.
\newblock {\em arXiv preprint arXiv:1908.07490}, 2019.

\bibitem{GFZSL}
Vinay~Kumar Verma and Piyush Rai.
\newblock A simple exponential family framework for zero-shot learning.
\newblock In {\em The European conference on machine learning and knowledge
  discovery in databases (ECML-PKDD)}, 2017.

\bibitem{wang2016learning}
Liwei Wang, Yin Li, and Svetlana Lazebnik.
\newblock Learning deep structure-preserving image-text embeddings.
\newblock In {\em Proceedings of the IEEE conference on computer vision and
  pattern recognition}, pages 5005--5013, 2016.

\bibitem{CUB}
P. Welinder, S. Branson, T. Mita, C. Wah, F. Schroff, S. Belongie, and P.
  Perona.
\newblock {Caltech-UCSD Birds 200}.
\newblock Technical Report CNS-TR-2010-001, California Institute of Technology,
  2010.

\bibitem{welinder2010caltech}
Peter Welinder, Steve Branson, Takeshi Mita, Catherine Wah, Florian Schroff,
  Serge Belongie, and Pietro Perona.
\newblock Caltech-ucsd birds 200.
\newblock 2010.

\bibitem{XianCVPR16}
Yongqin Xian, Zeynep Akata, Gaurav Sharma, Quynh Nguyen, Matthias Hein, and
  Bernt Schiele.
\newblock Latent embeddings for zero-shot classification.
\newblock In {\em IEEE Conference on Computer Vision and Pattern Recognition
  (CVPR)}, 2016.

\bibitem{xian2018tPAMI}
Yongqin Xian, Christoph~H Lampert, Bernt Schiele, and Zeynep Akata.
\newblock Zero-shot learning-a comprehensive evaluation of the good, the bad
  and the ugly.
\newblock {\em The IEEE Transactions on Pattern Analysis and Machine
  Intelligence}, 2018.

\bibitem{xian2018zero}
Yongqin Xian, Christoph~H Lampert, Bernt Schiele, and Zeynep Akata.
\newblock Zero-shot learning—a comprehensive evaluation of the good, the bad
  and the ugly.
\newblock {\em IEEE transactions on pattern analysis and machine intelligence},
  41(9):2251--2265, 2018.

\bibitem{xian2018feature}
Yongqin Xian, Tobias Lorenz, Bernt Schiele, and Zeynep Akata.
\newblock Feature generating networks for zero-shot learning.
\newblock In {\em Proceedings of the IEEE conference on computer vision and
  pattern recognition}, pages 5542--5551, 2018.

\bibitem{Xian_2018_CVPR}
Yongqin Xian, Tobias Lorenz, Bernt Schiele, and Zeynep Akata.
\newblock Feature generating networks for zero-shot learning.
\newblock In {\em The IEEE Conference on Computer Vision and Pattern
  Recognition (CVPR)}, June 2018.

\bibitem{Xian_2019_CVPR}
Yongqin Xian, Saurabh Sharma, Bernt Schiele, and Zeynep Akata.
\newblock {F-VAEGAN-D2: A Feature Generating Framework for Any-Shot Learning}.
\newblock In {\em The IEEE Conference on Computer Vision and Pattern
  Recognition (CVPR)}, June 2019.

\bibitem{Xian19}
Yongqin Xian, Saurabh Sharma, Bernt Schiele, and Zeynep Akata.
\newblock {f-VAEGAN-D2}: A feature generating framework for any-shot learning.
\newblock In {\em IEEE Conference on Computer Vision and Pattern Recognition
  (CVPR)}, 2019.

\bibitem{ye2017zero}
Meng Ye and Yuhong Guo.
\newblock Zero-shot classification with discriminative semantic representation
  learning.
\newblock In {\em Proceedings of the IEEE Conference on Computer Vision and
  Pattern Recognition}, pages 7140--7148, 2017.

\bibitem{APY}
Y. Yu.
\newblock {aPascal-aYahoo Image Data Collection}.
\newblock Technical report, University of Illinois at Urbana-Champaign, 2009.

\bibitem{Yu20}
Yunlong Yu, Zhong Ji, Jungong Han, and Zhongfei Zhang.
\newblock Episode-based prototype generating network for zero-shot learning.
\newblock In {\em IEEE Conference on Computer Vision and Pattern Recognition
  (CVPR)}, 2020.

\bibitem{zhang2018zero}
Hongguang Zhang and Piotr Koniusz.
\newblock Zero-shot kernel learning.
\newblock In {\em The IEEE Conference on Computer Vision and Pattern
  Recognition (CVPR)}, pages 7670--7679, 2018.

\bibitem{Zhang18}
Hongguang Zhang and Piotr Koniusz.
\newblock Zero-shot kernel learning.
\newblock In {\em IEEE Conference on Computer Vision and Pattern Recognition
  (CVPR)}, 2018.

\bibitem{Zhu:ICCV19}
Yizhe Zhu, Jianwen Xie, Bingchen Liu, and Ahmed Elgammal.
\newblock Learning feature-to-feature translator by alternating
  back-propagation for generative zero-shot learning.
\newblock In {\em The IEEE International Conference on Computer Vision (ICCV)},
  2019.

\end{thebibliography}
}

\end{document}